\documentclass[times,twocolumn,final,5p,hyperref={colorlinks=true, breaklinks}]{elsarticle}
\usepackage{stfloats}
\usepackage{float}
\usepackage{graphicx}
\usepackage{flushend}
\usepackage{prletters}
\usepackage{amsfonts,amsmath,amssymb,amsthm}
\usepackage{bm}
\usepackage{url}
\usepackage{comment}
\usepackage{caption}
\usepackage{graphicx}
\usepackage{float} 
\usepackage{subcaption}
\usepackage{enumerate}
\usepackage[utf8]{inputenc}
\usepackage{mathtools}
\usepackage{multirow,tabularx}
\usepackage{paralist}
\usepackage{booktabs}
\usepackage[disable]{todonotes}
\usepackage[font=small,labelfont=bf,labelsep=none]{caption}
\usepackage{tikz,xcolor,hyperref}
\definecolor{lime}{HTML}{A6CE39}
\usepackage{amssymb}
\usepackage{latexsym}
\usepackage{url}
\usepackage{xcolor}
\definecolor{newcolor}{rgb}{.8,.349,.1}
\newif\ifplnote
\plnotefalse
\definecolor{Set1g}{HTML}{1b9e77}
\definecolor{Set1b}{HTML}{377eb8}
\definecolor{Set1c}{HTML}{277e77}
\journal{Pattern Recognition Letters}

\begin{document}

\title{A Stronger Stitching Algorithm for Fisheye Images based on Deblurring and Registration}

\author[1]{Jing \snm{Hao}$^\dag$}\ead{isjinghao@gmail.com}
\author[1]{Jingming  \snm{Xie}$^\dag$} 
\author[1]{Jinyuan \snm{Zhang}}
\author[1]{Moyun \snm{Liu}}
\address[1]{School of Mechanical Science and Engineering, Huazhong University of Science and Technology, Hubei, China}
\nonumnote{$\dag$: Equal Contribution. }

\begin{abstract}
	Fisheye lens, which is suitable for panoramic imaging, has the prominent advantage of a large field of view and low cost. However, the fisheye image has a severe geometric distortion which may interfere with the stage of image registration and stitching. Aiming to resolve this drawback, we devise a stronger stitching algorithm for fisheye images by combining the traditional image processing method with deep learning. In the stage of fisheye image correction, we propose the Attention-based Nonlinear Activation Free Network (ANAFNet) to deblur fisheye images corrected by Zhang’s calibration method. Specifically, ANAFNet adopts the classical single-stage U-shaped architecture based on convolutional neural networks with soft-attention technique and it can restore a sharp image from a blurred image effectively. In the part of image registration, we propose the ORB-FREAK-GMS (OFG), a comprehensive image matching algorithm, to improve the accuracy of image registration. Experimental results demonstrate that panoramic images of superior quality stitching by fisheye images can be obtained through our method.

\end{abstract}

\begin{keyword}
	image stitching 
    \sep image deblurring \sep image registration \sep deep learning.
\end{keyword}

\maketitle

\markboth{TODO}%
{Hao \MakeLowercase{\textit{et al.}}: A Stronger Stitching Algorithm for Fisheye Images based on Deblurring and Registration}

\section{Introduction} \label{sec:introduction}

Panoramic vision has been widely applied in security monitoring, driving assistance, drone aerial photography, and remote sensing images. Characterized by its short focal length and large field of view, the fisheye lens is suitable for panoramic imaging tasks. In generating panoramic images with the same field of view, a fisheye camera can reduce the number of sub-images used for stitching and decrease time consumption as well as system design costs. Image stitching can be accomplished through such steps as image preprocessing, image registration, and image blending \cite{1}, among which the significance of accurate registration remains unquestionable. Based on image registration, image stitching can be grouped into area-based image stitching and feature-based image stitching. As for area-based stitching algorithm, it relies on the similarity of the overlapping regions between “windows” of pixel values in the two images which need to be stitched. Therefore, it has a strict limitation on overlapping areas. While the feature-based algorithm computes the correspondence in multiple images based on the descriptors generated by low-level features including edge, corner, color, and histogram, it is more robust to affine transformation and projection transformation than the area-based image stitching method. Due to severe radial distortion in the fisheye images, we choose the automatic panoramic image stitching method belonging to the feature-based algorithm proposed by \cite{2} as the baseline of image stitching. This method formulates stitching as a multi-image matching problem, hence it is insensitive to ordering, orientation, scale, and illumination in the process of image registration. 

Zhang’s calibration method \cite{3} is firstly applied to correct the fisheye images because of the severe distortion in the fisheye image. However, the stretch blurring still exists in the image after correction. This deformation would significantly impede the subsequent image registration process, and will ultimately lower the quality of the stitched image. Image deblurring \cite{4} can solve this problem because it aims to transform the blurred image into a sharp one. Recent technological advances in deep learning have revolutionized the field of image deblurring. Several methods based on neural networks have advanced the state of the art compared with the traditional image deblurring algorithm. The data-driven approach to image deblurring can fit arbitrary deblurring mapping functions with given training data, which is superior to the traditional method that makes use of some prior information in the natural image to restore images.

Nonlinear Activation Free Network \cite{5}, an image restoration neural network, is applied to deblur images and the attention mechanism \cite{6} is introduced to improve the ability of deblurring. Moreover, it is obvious that the quality of panoramic images directly depends on the accuracy of image registration to a considerable extent, we propose a new image registration scheme aiming to improve the matching accuracy of feature points. Experimental results have proved that the quality of panoramic image stitching based on fisheye images can be effectively improved in both visual and quantitative metrics.

The remainder of this paper is organized as follows: Section~\ref{sec: methods} demonstrates the principles of proposed deblurring and registration algorithm, as well as the pipeline of image stitching. Section~\ref{sec: experiments} verifies the effectiveness of the proposed algorithms and the gains in image stitching through a series of experiments. Section~\ref{sec: conclusions} presents the summary and future scope.

\section{The proposed method} \label{sec: methods}
The quality of fisheye image distortion correction directly determines the accuracy of the image registration relationship. In this paper, the fisheye image is corrected by Zhang's calibration method \cite{3}, and an image deblurring neural network is proposed to solve the problem of blurring artifacts of the corrected image. In order to improve the image registration accuracy, we propose the ORB-FREAK-GMS (OFG) image matching algorithm. Finally, the image stitching pipeline is briefly introduced.

\subsection{Image deblurring}

With the emergence of deep learning, the performance of image deblurring has improved significantly. The image deblurring method based on deep learning can fit any fuzzy distortion in the image through the neural network, and restore the image with high quality. The Attention-based Nonlinear Activation Free Network (ANAFNet), which introduced the attention gate to NAFNet \cite{5} is proposed to enhance the model sensitivity and image restoration quality. ANAFNet adopts the classical single-stage U-shaped architecture with the soft-attention technique, and its architecture is shown in Fig. 1. NAFNet, as an image restoration network, replaced the nonlinear activation functions with SimpleGate and achieved excellent results on image restoration tasks with lower computational costs. NAFNet consists of encoders and decoders, all of which followed the same block. Each of the blocks consists of Layer Normalization \cite{6}, convolution and its variants \cite{7}, SimpleGate, and Simple Channel Attention (SCA). SimpleGate is a way to replace GELU activation \cite{9} without loss of performance. It is implemented by directly dividing the feature map into two parts in the channel dimension and multiplying them. SCA simplified the SE module \cite{8} by reserving only one 1×1 convolution and the rest of the components(other convolutions and activations) are removed considering the model complexity.

\begin{figure}
\centerline{\includegraphics[width=\columnwidth]{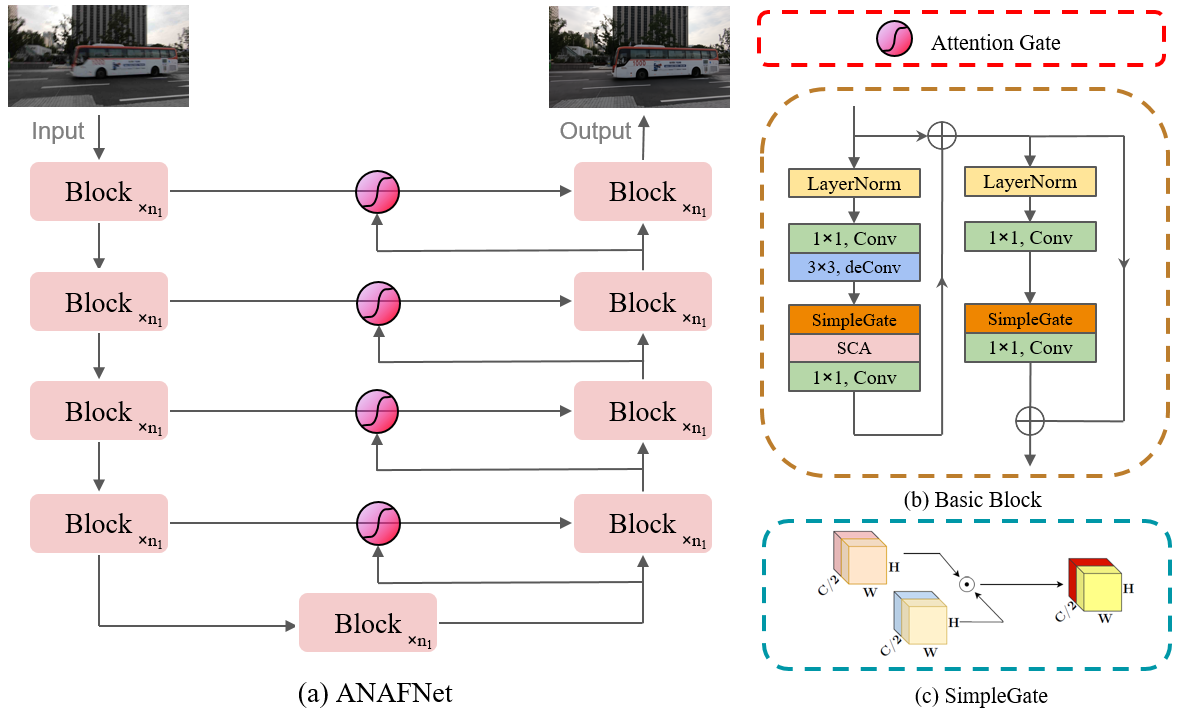}}
\caption{ The ANAFNet overview. It is adopts the U-shaped architecture with multiple basic block as whole framework. The basic block consists of convolutions, SimpleGate and simple channel attention(SCA). Furthermore, the attention gate is added to skip-connections. $\bigoplus$/$\bigodot$: element-wise add/multiplication.}
\end{figure}

NAFNet utilizes skip-connections to aggregate the features which come from encoders and their corresponding decoders. While this kind of hard element-wise fusion may disturb the feature distribution and give rise to inconsistency between encoders and decoders. Besides, NAFNet only uses the attention mechanism within inner blocks without taking attention-based skip-connections into consideration. Hence, the attention gate [10], a simple soft attention mechanism, is introduced to capture potential correlations between encoders and decoders. It provides an adaptive way to aggregate features that come from different blocks within the minimal computational overhead. The architecture of the attention gate is shown in Fig. 2, which can be formulated as follows:

\begin{figure}[htbp]
\centerline{\includegraphics[width=\columnwidth]{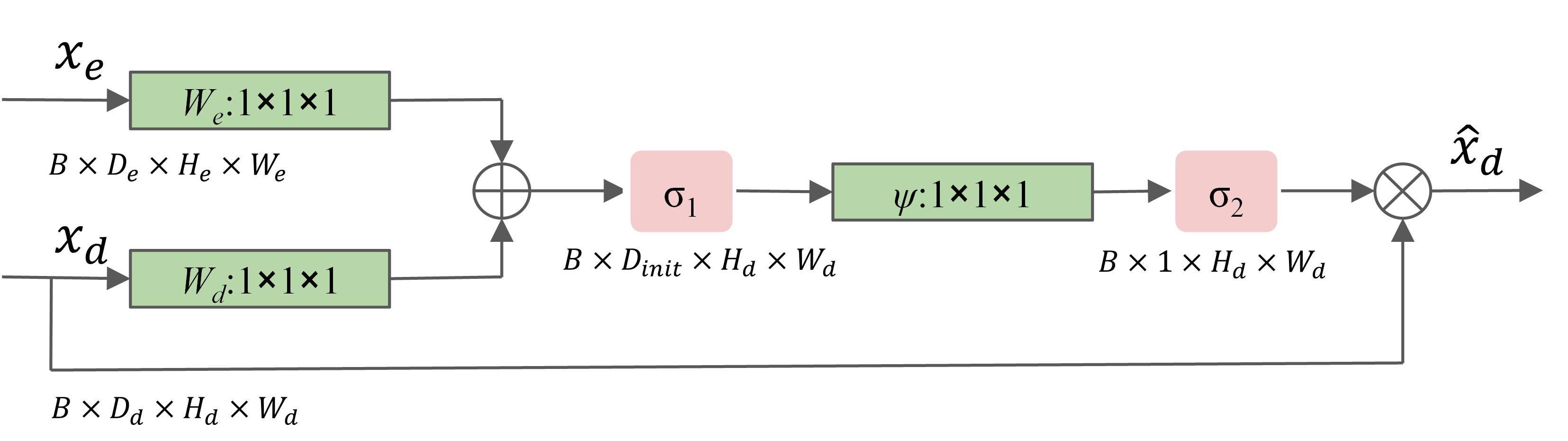}}
\caption{\ The architecture of attention gate.}
\end{figure}
$$ W_{attn} = \psi^{T} (\sigma_{1}( W_{e}^{T}x_{e} +  W_{d}^{T}x_{d} + b_{e} ))+b_{\psi},$$
$$\hat{x_{d}} =\sigma_{2}(W_{attn})*x_{d},$$
where $\sigma_{1}$ and $\sigma_{2}$ correspond to activation function. Attention gate is characterized by a set of parameters, including linear transformations $ W_{e}\in\mathbf{R}^{D_{einit}} $, $ W_{d}\in\mathbf{R}^{D_{dinit}} $, $ \psi\in\mathbf{R}^{D_{init}} $ and bias terms $ b_{g}\in\mathbf{R}^{D_{init}} $, $ b_{\psi}\in\mathbf{R} $.

The linear transformations are computed by using channel-wise 1×1×1 convolutions for the input tensors. As the NATNet replaces the activation with SimpleGate module, $\sigma_{1}$ and $\sigma_{2}$ are directly set as the identity function to maintain the attention gate within a low computational cost. The output of the attention gate is the element-wise multiplication of input feature maps and attention coefficients.

\subsection{Image registration}
The image registration makes up of feature point detection and feature point matching. We propose the OFG method, a new image registration scheme that can improve the matching accuracy of feature points effectively. Feature point detection is composed of keypoint detection and descriptor generation. Considering that ORB \cite{11} is two orders of magnitude faster than sift \cite{12} and one order than surf \cite{13}, ORB is chosen as the basic feature point detector \cite{14}. In terms of the descriptor generation, ORB generates n-bit binary vectors with rotation invariance by BRIEF \cite{15} which utilizes the random pair sampling to compare pairs of pixel intensities. Intuitively, visual prior information can be introduced to describe keypoints more comprehensively. Propelled by the human visual system and exactly the retina, FREAK \cite{16} uses the retinal sampling grid which has the highest density of points near the center while the density diminishes gradually as moving far from the keypoints. Through overlapping receptive fields, a more unique and distinguishable descriptor can be obtained. Therefore, FREAK is utilized to generate descriptors that are needed in the subsequent matching step.

In the process of image matching, the brute force method \cite{17} is widely used to match the feature points, but the mismatch rate is pretty high. RANSAC \cite{18} is applied to filter the matching outliers. Owing to its inconsideration of the 3D location distribution information, the incorrectly matched pairs still exist. Grid-based Motion Statistics (GMS) \cite{19} is a statistical formulation for partitioning true and false matches based on the number of neighboring matches. It calculates the score of each neighborhood that contains feature points and it is a real-time feature matcher because of the grid-based score estimator. Therefore, GMS is chosen as the keypoint matching method instead of RANSAC. We note the ORB-FREAK-GMS comprehensive image registration as OFG algorithm, which will be used in the feature point-based image mosaicking pipeline.

\begin{figure}[htbp]
	\centering
	\begin{subfigure}{0.49\linewidth}
	    \centering
	    \includegraphics[width=0.9\linewidth]{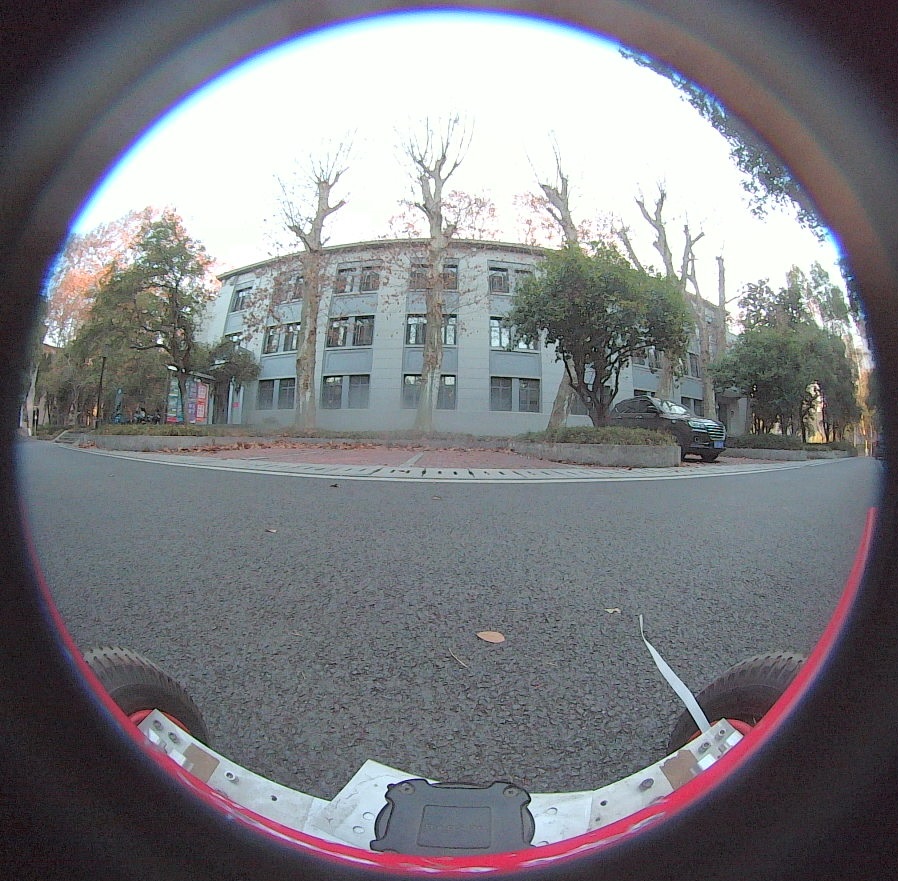}
		\caption{Fisheye Image}
	\end{subfigure}
	\begin{subfigure}{0.49\linewidth}
	    \centering
	    \includegraphics[width=0.9\linewidth]{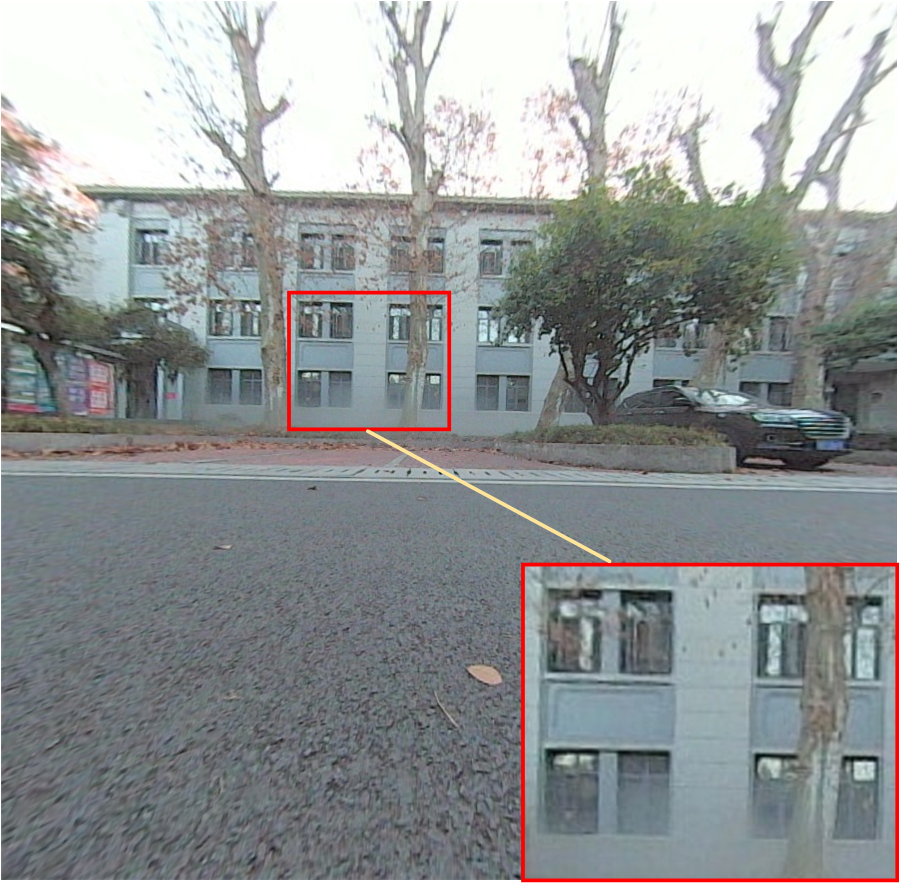}
		\caption{Zhang's correction}
	\end{subfigure}
	\begin{subfigure}{0.49\linewidth}
	    \centering
	    \includegraphics[width=0.9\linewidth]{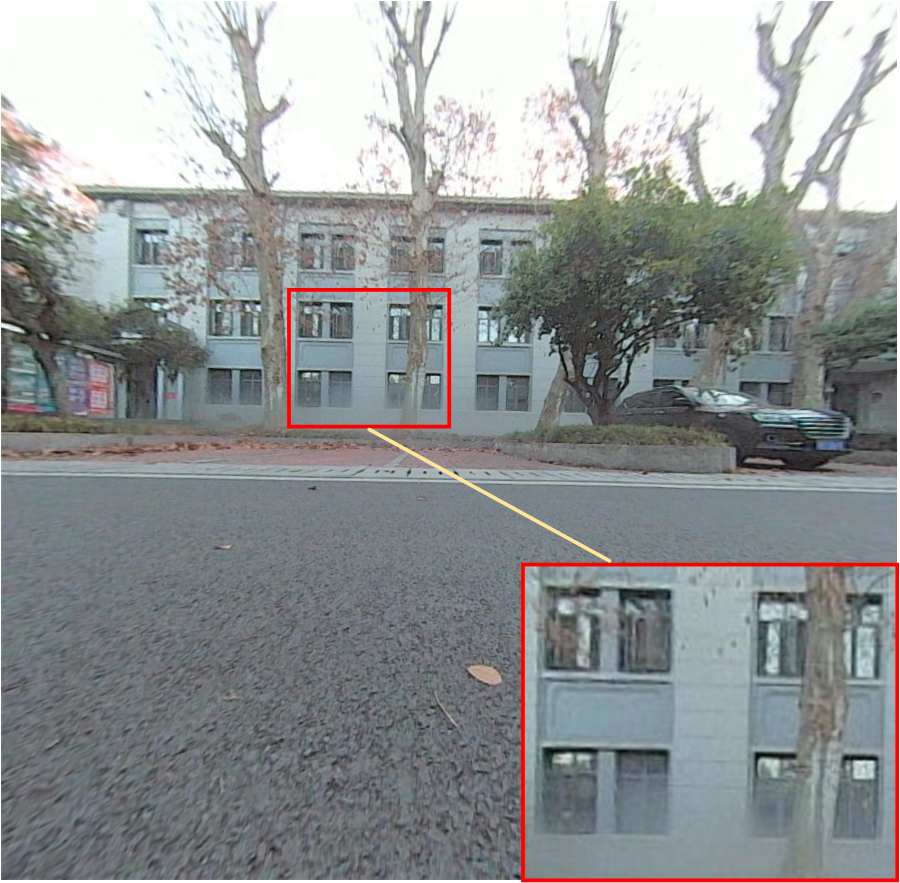}
		\caption{Guided Filtering}
	\end{subfigure}
	\begin{subfigure}{0.49\linewidth}
	    \centering
		\includegraphics[width=0.9\linewidth]{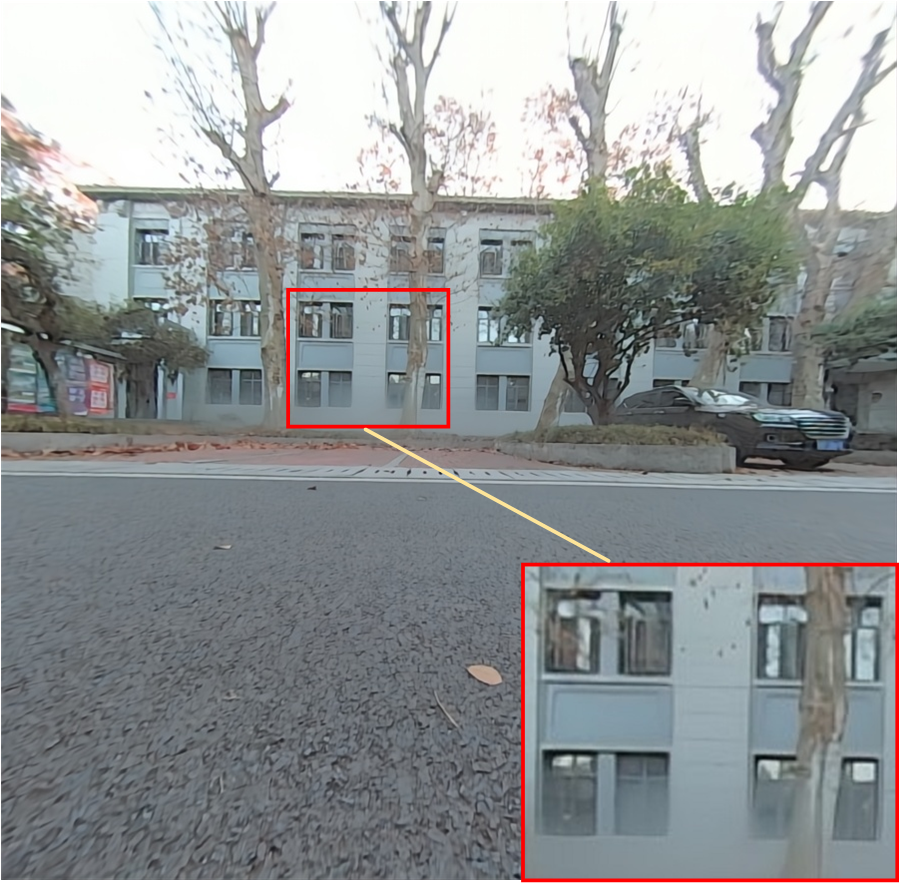}
		\caption{ANAFNet}
	\end{subfigure}
    \caption{\ The visual comparison of image deblurring method.}
\end{figure}
\vspace{-0.3cm}

\subsection{Image stitching pipeline}
We follow \cite{2} to conduct automated panoramic image stitching. Firstly, the fisheye images are corrected by Zhang’s calibration method and deblurred by ANAFNet. After that, the OFG algorithm is used to detect feature points and match them effectively and efficiently. Then the bundle adjustment is adopted to solve for all of the camera parameters jointly. Gain compensation is introduced to eliminate the brightness difference caused by different illumination. Next, the stitching seam lying in the overlapping area of adjacent images are obtained by seam searching method in order to removing the ghost effect. Finally the multi-band algorithm is utilized to fuse multiple images to generate panoramic images.

\section{Experiments and discussion} \label{sec: experiments}
\subsection{Distortion correction}
\subsubsection{Dataset and training details}
REDS dataset \cite{20} is employed in this paper. It consists of 300 video sequences having a length of 100 frames with 720×1280 resolution. The frames are of high quality in terms of the reference frames, diverse scenes and locations, realistic approximation of blurs, and the used standard lossy compression artifacts. REDS covers a large diversity of contents, people, handmade objects, and environments that can bring great benefits in terms of model generalizability. The pair of training datasets and ground-truth is composed of blurry and sharp frames. We train ANAFNet with AdamW \cite{21} optimizer ($\beta_{1}$ = 0.9, $\beta_{2}$ = 0.9, weight decay 1e-3) for total 400K iterations with the initial learning rate 5e-4 gradually reduced to 1e-7 with the cosine annealing schedule \cite{22}. We train models on 256×256 patches with mini-batchsize of 8 utilizing 4×11G-GTX 2080Ti GPUs, and we use PSNR loss \cite{23} as the loss function and apply flip and rotation as data augmentation.
\begin{table}[htb]
    \renewcommand\arraystretch{1.1}
    \centering
    \caption{Image quality evaluation on deblurring methods.}
    \vspace{-0.2cm}
    {
    \resizebox{\linewidth}{!}{
        \begin{tabular}{ccccc}
        \toprule
            \multicolumn{2}{c}{}&\multicolumn{1}{c}{BG}&\multicolumn{1}{c}{entropy}&\multicolumn{1}{c}{contrast} \\
            \hline
            \multicolumn{1}{c}{\multirow{4}{*}{Outdoor}}&Zhang’s calibration\cite{3}&223.353&7.0365&97\\
            \multicolumn{1}{c}{}&Guided Filtering\cite{24}&233.951&6.86489&102\\
            \multicolumn{1}{c}{}&NAFNet\cite{5}&257.058&6.98638&101\\
            \multicolumn{1}{c}{}& Ours & \textbf{262.815} & \textbf{7.01984} & \textbf{105}  \\
            \hline
            \multicolumn{1}{c}{\multirow{4}{*}{Indoor}}&Zhang’s calibration\cite{3}& 97.758 & 7.49556 & 56\\
            \multicolumn{1}{c}{}&Guided Filtering\cite{24}& 98.075 & 7.47419 & 56\\
            \multicolumn{1}{c}{}&NAFNet\cite{5}& 105.564 & 7.47439 & 56\\
            \multicolumn{1}{c}{}& Ours & \textbf{111.953} & \textbf{7.48100} &\textbf{59}\\
        \bottomrule
        \end{tabular}
        }
        \label{tab:1}
    }
\end{table}
\subsubsection{Evaluation and analysis}

To evaluate the image deblurring method, fisheye images are firstly corrected through Zhang’s calibration method, and the images are deblurred after correction. The performance of several image deblurring methods is evaluated in terms of visual quality and quantitative metrics: 

1) Qualitative results. Our method achieves better qualitative results compared with the traditional deblurring algorithm. The qualitative comparisons are shown in Fig. 3, in which Fig. 3(a) is an image captured by a fisheye camera. There is an obvious phenomenon that existing stretch blur deformation in the radial direction. Fig. 3(b) is the image corrected by Zhang's calibration method while the stretch blur and noise still present in the rectified image. Fig. 3(c) is the image deblurred by guided filtering \cite{24} that is a famous edge-preserving smoothing algorithm and Fig. 3(d) is the image deblurred by ANAFNet. It is apparent that Fig. 3(d) is visually superior to Fig. 3(c) which is generated by traditional deblurring method. \\
\begin{table*}[bp]
    \centering
    \caption{\qquad \quad The Correct matching rate \% of different image registration methods on Mikolajczyk \cite{26} .}
    \setlength{\tabcolsep}{5mm}
    \begin{tabular}{cccccccc}
    \toprule
    \multicolumn{1}{c}{Dataset}&\multicolumn{1}{c}{Method}&\multicolumn{1}{c}{1-2}&\multicolumn{1}{c}{1-3}&\multicolumn{1}{c}{1-4}&\multicolumn{1}{c}{1-5}&\multicolumn{1}{c}{1-6}&\multicolumn{1}{c}{Average}\\
    \hline
    \multicolumn{1}{c}{\multirow{3}{*}{Ubc}}& ORB+RNASAC & 98.54 & 98.08 & 96.22 & 93.93 & 88.78 & \textbf{95.11}   \\
    \multicolumn{1}{c}{}& ORB+GMS & 98.91 & 98.36 & 97.11 & 95.00 & 90.64 & \textbf{96.00}   \\
    \multicolumn{1}{c}{}& ours & 99.34 & 99.27 & 98.52 & 97.45 & 95.02 & \textbf{97.92}\\
    \hline
    \multicolumn{1}{c}{\multirow{3}{*}{Graf}}&ORB+RNASAC & 91.14 & 75.01 & 53.30 & 8.46 & 0 & \textbf{45.58 }  \\
    \multicolumn{1}{c}{}&ORB+GMS & 94.30 & 82.94 & 71.82 & 0 & 0 & \textbf{49.81 } \\ 
    \multicolumn{1}{c}{}&ours & 97.26 & 86.66 & 75.00 & 74.47 & 46.34 & \textbf{75.95}   \\ 
    \hline
    \multicolumn{1}{c}{\multirow{3}{*}{Leuven}}&ORB+RNASAC & 92.03 & 91.32 & 90.39 & 87.73 & 86.17 & \textbf{89.53}   \\
    \multicolumn{1}{c}{}& ORB+GMS & 93.43 & 92.33 & 92.04 & 90.07 & 88.45 & \textbf{91.26}   \\ 
    \multicolumn{1}{c}{} & ours & 96.88 & 95.43 & 93.27 & 92.65 & 90.38 & \textbf{93.72}   \\ 
    \hline
    \multicolumn{1}{c}{\multirow{3}{*}{Bikes}}& ORB+RNASAC & 95.20 & 92.99 & 91.03 & 86.98 & 77.40 & \textbf{88.72}   \\ 
    \multicolumn{1}{c}{}& ORB+GMS & 96.21 & 95.09 & 92.58 & 89.61 & 81.70 & \textbf{91.04}   \\
    \multicolumn{1}{c}{}& ours & 98.11 & 97.58 & 96.66 & 94.36 & 90.85 & \textbf{95.51}   \\
    \bottomrule
    \end{tabular}
    \label{tab:my_label}
\end{table*}
2) Quantitative results. Table \ref{tab:1} shows the quantitative results of several deblurring methods. Three non-reference image quality metrics including Brenner gradient(BG), entropy and contrast \cite{25} are used to make an objective evaluation. As shown in Table \ref{tab:1}, outdoor and indoor scenes are selected for experiments in order to guarantee the consistency in the different situations. It can be seen that ANAFNet improves deblurring effect excellently, and the deblurring method based on neural network outperforms the one based on statistics. As for outdoor and indoor images, the Brenner gradient increases 17.67\% and 14.52\%, respectively. The contrast increases 4.12\% and 5.36\%, respectively. The original blurry image gets the highest value on the entropy metric while there is an obvious stretch blur artifacts shown in Fig. 3(b). Hence we conjecture that blur noise could fluctuate the entropy metric. Furthermore, ANAFNet method gets the highest value on entropy compared with other deblurring methods, being in line with the changing trend of Brenner gradient and contrast.

\begin{figure*}[ht]
	\centering
	\begin{subfigure}{1\linewidth}
	    \centering
	    \includegraphics[width=0.7\linewidth]{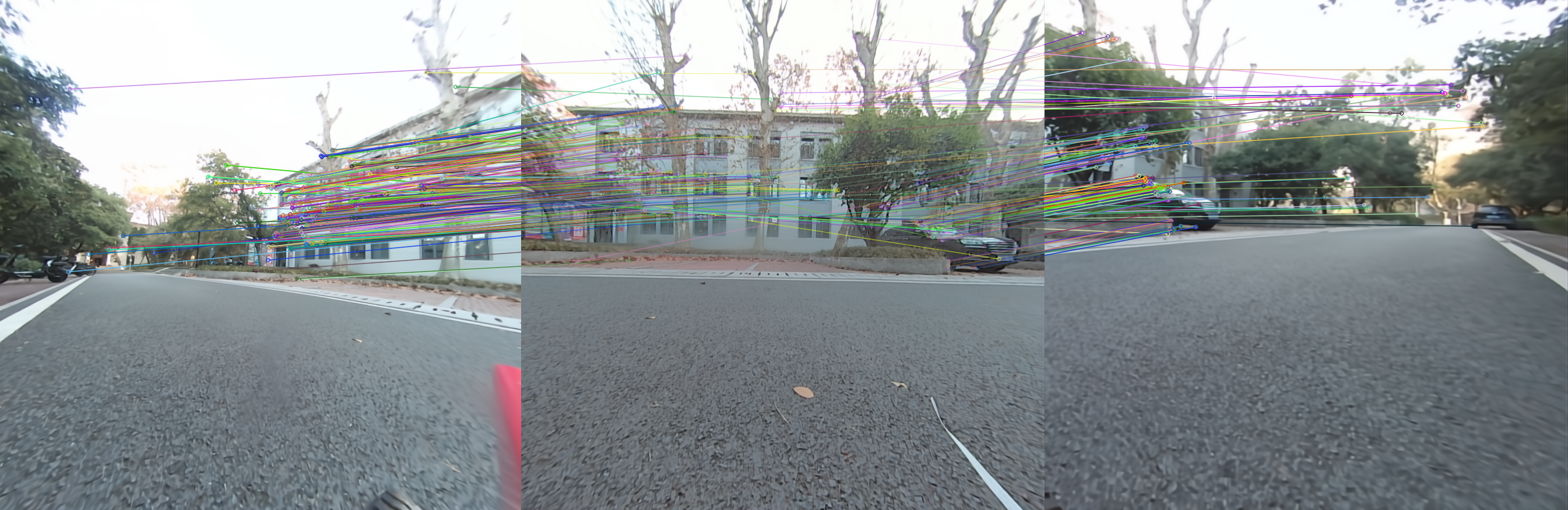}
		\caption{The feature point matching result by approach [2].}
	\end{subfigure}
	\begin{subfigure}{1\linewidth}
	    \centering
	    \includegraphics[width=0.7\linewidth]{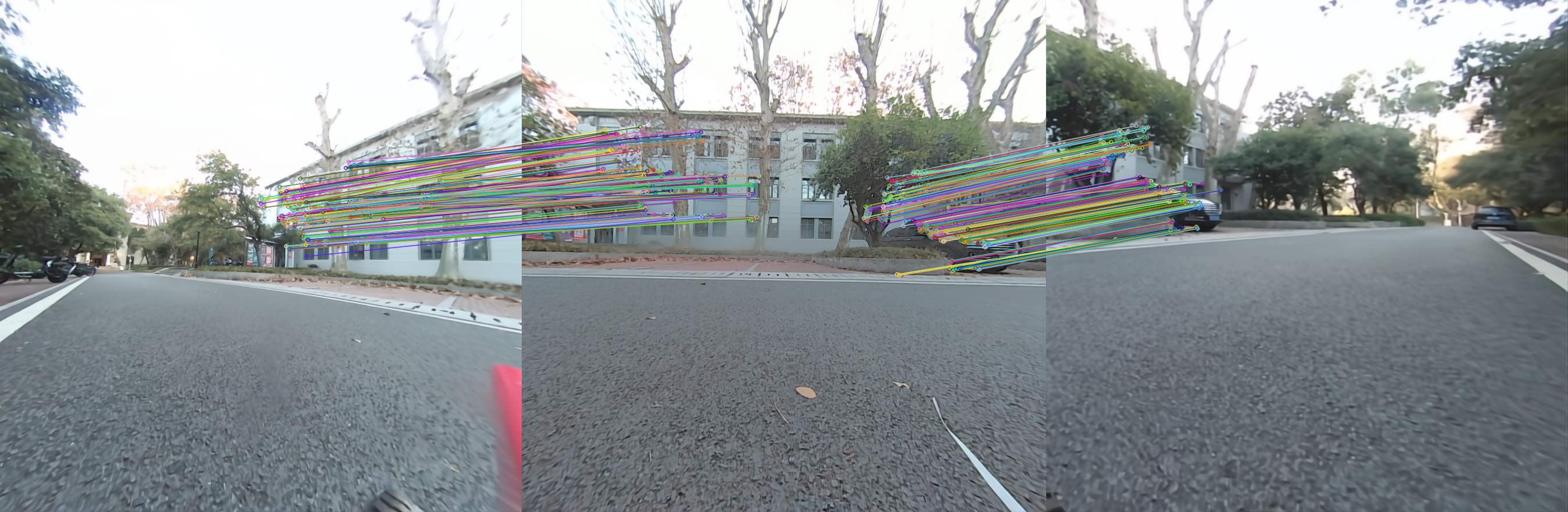}
		\caption{The feature point matching result by ours.}
	\end{subfigure}
    \caption{\ Feature point matching result.}
\end{figure*}

\begin{figure*}[ht]
\centerline{\includegraphics[width=0.7\linewidth]{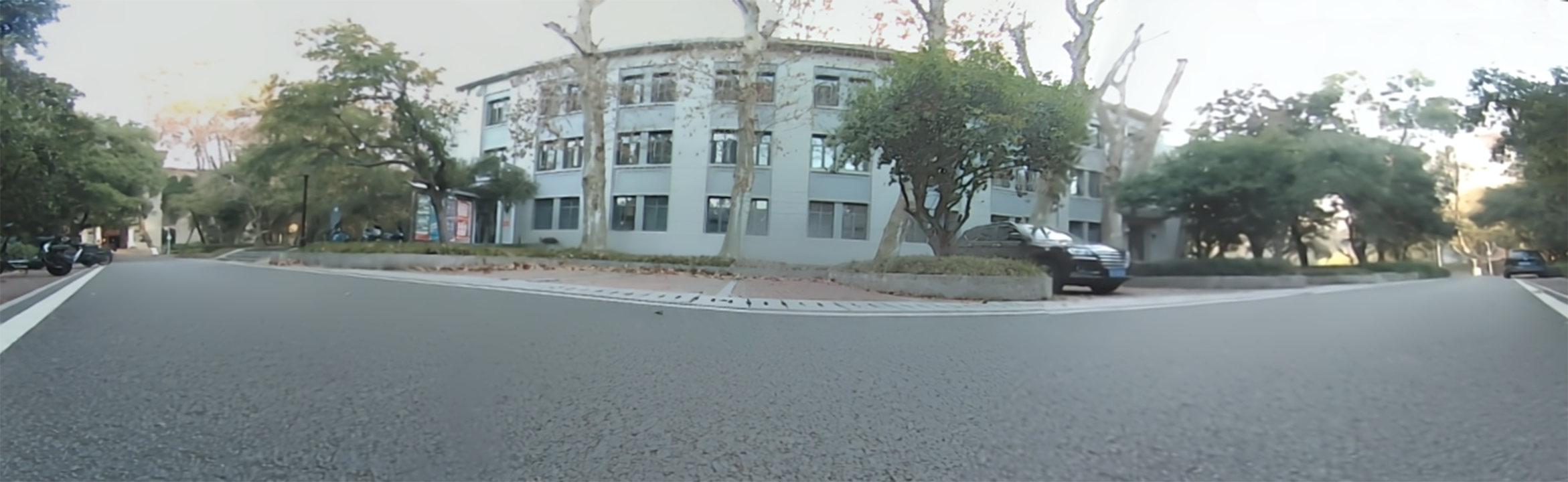}}
\caption{\ The stitched result on outdoor images by ours.}
\end{figure*}
\subsection{Image registration}
The OFG algorithm is evaluated on Mikolajczyk \cite{26} public dataset which covers a variety of transformations such as scale, illumination, rotation and JPEG compression. Each set of image sequences contains a base image and five deformed images. The degree of image deformation increases gradually.

Mikolajczyk provides a true homography matrix of image transformations, which can be used to describe the transformation relationship among feature points. Four sets of image sequences covering multiple transformations are selected so that the effectiveness and robustness of algorithm can be verified from multiple perspectives.\par
The matching accuracy $A_{match}$ is formulated as  $A_{match}=correctMatches/Matches * 100\%$, where $Matches$ represents the number of matching pair filtered by different matching method. It is noted that if the Euclidean distance between the transformed point by homography matrix and its corresponding matching point is less than the distance threshold, the matching pair is regarded as one correct matching pair. In the experiment, the number of feature points is 5000 and the distance threshold is 5 pixels.\par
As shown in Table \ref{tab:my_label}, the OFG method achieves highest average  matching accuracy among four image sequences. In terms of rotation, scale, illumination and blur transformation, the average matching accuracy of OFG method increase by 2.81\%, 30.37\%, 4.19\% and 6.79\% respectively compared with ORB+RANSAC on four public datasets. And as for descriptor generation method, the matching accuracy of FREAK is 1.92\%, 26.14\%, 2.46\% and 2.32\% higher than that of BRIEF, respectively. For the graf sequence, with the increase of deformation, the matching accuracy of the traditional ORB algorithm decreases to zero. It is a fact that the Euclidean distance between the matching pairs has been greater than 5 pixels, a large deviation of correct position. However, our OFG algorithm can still achieve 46.34\% matching accuracy, which verifies its superiority. Generally, the OFG algorithm shows extremely high matching accuracy and robustness in four transformations on Mikolajczyk, which provides a more accurate image matching relationship for image stitching.

\subsection{Panoramic imaging}
We follow the image stitching pipeline proposed by M. Brown \cite{2}, and introduce the ANAFNet as well as OFG algorithm to improve the quality of stitched images. In outdoor images, there is a huge difference in brightness in different directions. Object shape and surface texture also vary greatly. Moreover, there is usually more noise in the image at locations such as the ground, sky, and leaves which may interfere with the image registration. Considering that, stitching outdoor images tends to be more difficult than indoor images. Therefore, we perform stitching experiments on outdoor images and the visual results of the outdoor panoramic image are provided.
In the process of stitching three outdoor images, it can be found that method \cite{2} cannot stitch them successfully, but our method can do it well. Fig. 4 shows the feature point matching relationship of the sub-images. It can be seen that there is a large number of wrong feature point matching results by approach [2] (the RANSAC confidence is set to 0.99 in the experiment) in Fig. 4(a) which will impede the subsequent stage of image registration and stitching. Fig. 4(b) shows that our method can ensure the correctness of the feature point matching relationship between the sub-images. Fig. 5 shows the stitching result by our method. It can be found that there is a huge degree of scale and rotation transformation between the stitched sub-images in Fig. 4(b), hence the robustness of the proposed method to image noise, exposure, scale, and rotation transformations has been verified in the stitching task.

\section{Conclusion} \label{sec: conclusions}
In this letter, we propose a stronger stitching algorithm to generate panoramic images within fisheye images. First, we devise ANAFNet, an image deblurring neural network combined with soft attention mechanism, which eliminates the artifacts of the images corrected by Zhang’s calibration method. Then, we propose the OFG algorithm to improve the accuracy and robustness of image registration. In addition, extensive experiments on image deblurring and image registration validate the superiority of our proposed algorithm. A satisfying panoramic image stitching result is finally achieved no matter on visual quality or quantitative metrics. Our future research will focus on video stitching method.

\section*{Declaration of Competing Interest}
We have no conflict of interest to declare.







\end{document}